\documentclass{article}
\usepackage{spconf}
\usepackage{amsmath,graphicx}
\usepackage{booktabs}
\usepackage{stackengine}

\usepackage{bm}
\usepackage{multirow}
\usepackage{array}
\usepackage{caption}
\usepackage{subcaption}
\newcommand\Tstrut{\rule{0pt}{2.3ex}}
\usepackage{amssymb}
\usepackage{lipsum}

\usepackage{hyperref} 
\usepackage{xcolor}  

\title{End-to-end Whispered Speech Recognition\\with Frequency-weighted Approaches and Pseudo Whisper Pre-training}
\name{Heng-Jui Chang, Alexander H. Liu, Hung-yi Lee, Lin-shan Lee}
\address{College of Electrical Engineering and Computer Science, National Taiwan University}

\usepackage{tikz}
\newcommand\copyrighttext{%
  \scriptsize Copyright 2021 IEEE. Published in the 2021 IEEE Spoken Language Technology Workshop (SLT) (SLT 2021), scheduled for 19-22 January 2021 in Shenzen, China. Personal use of this material is permitted. However, permission to reprint/republish this material for advertising or promotional purposes or for creating new collective works for resale or redistribution to servers or lists, or to reuse any copyrighted component of this work in other works, must be obtained from the IEEE. Contact: Manager, Copyrights and Permissions / IEEE Service Center / 445 Hoes Lane / P.O. Box 1331 / Piscataway, NJ 08855-1331, USA. Telephone: + Intl. 908-562-3966.}
\renewcommand\copyrightnotice{%
\begin{tikzpicture}[remember picture,overlay]
\node[anchor=south,yshift=12pt] at (current page.south) {\fbox{\parbox{\dimexpr\textwidth-\fboxsep-\fboxrule\relax}{\copyrighttext}}};
\end{tikzpicture}%
}

\begin{document}

\maketitle

\begin{abstract}
    Whispering is an important mode of human speech, but no end-to-end recognition results for it were reported yet, probably due to the scarcity of available whispered speech data. 
    In this paper, we present several approaches for end-to-end (E2E) recognition of whispered speech considering the special characteristics of whispered speech and the scarcity of data.
    This includes a frequency-weighted SpecAugment policy and a frequency-divided CNN feature extractor for better capturing the high-frequency structures of whispered speech, and a layer-wise transfer learning approach to pre-train a model with normal or normal-to-whispered converted speech then fine-tune it with whispered speech to bridge the gap between whispered and normal speech.
    We achieve an overall relative reduction of 19.8\% in PER and 44.4\% in CER on a relatively small whispered TIMIT corpus.
    The results indicate as long as we have a good E2E model pre-trained on normal or pseudo-whispered speech, a relatively small set of whispered speech may suffice to obtain a reasonably good E2E whispered speech recognizer.
\end{abstract}


\copyrightnotice
\vspace{-13pt}

\section{Introduction}
	\label{sec:intro}
	Although less frequently used than normal speech, whispering is a basic mode of human speech used on special occasions such as interchanging confidential information, having conversations in meetings, theaters or libraries, or for patients with impaired glottises.
	Machine recognition of whispered speech is crucial yet extremely difficult due to its unique nature, such as no vocal cord vibrations \cite{Jovicic08-analcons, Lim10-wtimit}, lower speaking rates \cite{Jovicic08-analcons, Lee14-iwhisp}, lower energy \cite{Ito05-analwhsp, Ghaffarzadegan16-pswhsp, Grozdic17-invfilt}, an upward shift of formant frequencies \cite{Ito05-analwhsp, Ghaffarzadegan16-pswhsp}, and flatter spectra \cite{Ito05-analwhsp, Ghaffarzadegan16-pswhsp, Grozdic17-invfilt, Morris02-recons}.
	Automatic speech recognition (ASR) models trained on normal speech are thus inevitably degraded severely for whispered speech due to such mismatch \cite{Ghaffarzadegan16-pswhsp, Grozdic17-invfilt}.
	Various approaches have been used to overcome these difficulties, including model adaptation~\cite{Lim10-wtimit, Jou05-articulatory, Mathur12-param}, pseudo whisper features~\cite{Ghaffarzadegan16-pswhsp, Grozdic17-invfilt, Ghaffarzadegan17-dnnsmall}, non-audible murmur microphone (NAM) \cite{Yang12-murmur}, articulatory features \cite{Jou05-articulatory, Srinivasan19-articulatory, Cao16-acticulatory}, and visual cues \cite{Tao14-lipreading, Tran13-audiovis, Petridis18-visual}, achieving substantial improvements primarily based on the earlier very successful hidden Markov models (HMM) \cite{Rabiner89-hmm}.
	        \begin{figure}[t]
			\centering
			\includegraphics[width=\linewidth]{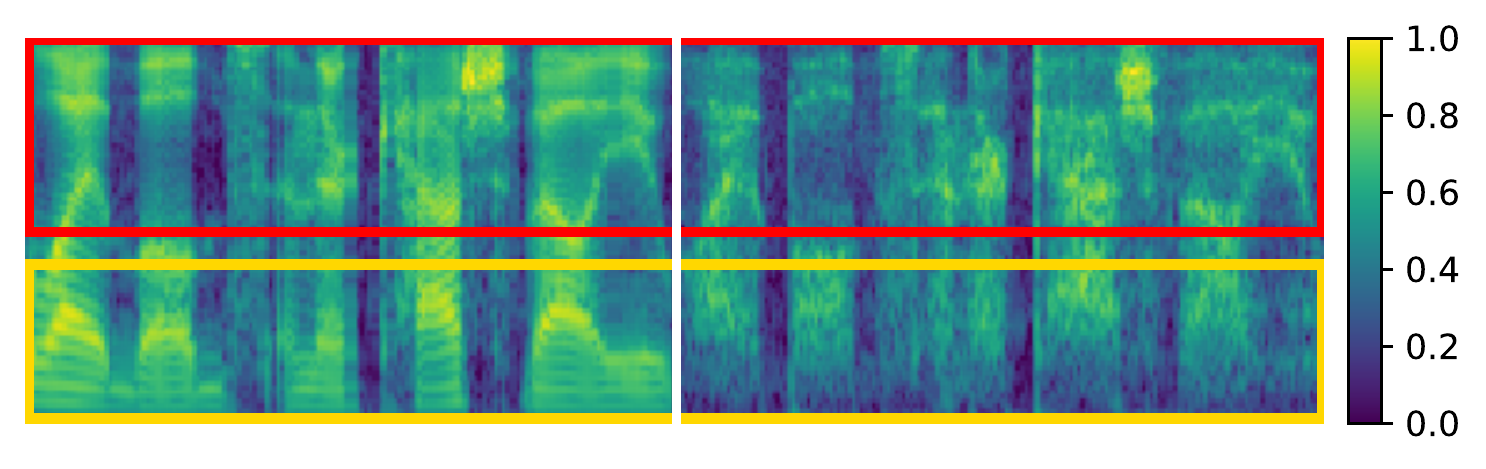}\\
			\vspace{-8pt}
            normal ~~~~~~~~~~~~~~~~~~~~~~~~~~~~~ whisper ~~~~~~~
            \vspace{-8pt}
			\caption{Mel-spectrograms of the same sentence produced by the same speaker in normal and whispered voice. 
			The high-frequency features (red box) are preserved in whispered speech, while the lower ones (yellow box) are seriously lost.}
			\label{fig:feats}
			\vspace{-6pt}
		\end{figure}

	Recently, E2E ASR approaches such as connectionist temporal classification (CTC) \cite{Graves14-ctcasr}, RNN-transducer \cite{Graves12-rnnt}, and Sequence-to-sequence model \cite{Chan16-las} have been overwhelmingly attractive and shown effective in globally optimizing the whole ASR process for the overall performance rather than locally optimizing acoustic and language models under different criteria.
	These approaches achieve fascinating accuracy as long as enough training data are available, without the need for hand-crafted modules or language-specific knowledge as in earlier approaches.

	However, the effectiveness of E2E approaches over whispered speech is yet to be confirmed.
	Previous works suggest deep learning useful for whispered ASR \cite{Markovic13-whispe, Grozdic13-appwhsp, Lim16-transfer}.
	Meanwhile, the success of E2E approaches for normal ASR is widely believed to depend on the quantity of data \cite{Amodei16-deep-speech-2, Chiu18-sotaasr} and the model architecture \cite{Amodei16-deep-speech-2, Hori17-jointctc, Zhang17-very-deep-conv, Han19-sotaasr}.
	Collecting whispered speech data of reasonable size is difficult, and the unique characteristics of whispered speech may need special considerations in model design and training.
	These are the questions this paper wishes to obtain at least some answers to.
    
	This paper is, to our knowledge, the earliest report focusing on whispered speech recognition with E2E models.
    We propose a frequency-weighted SpecAugment \cite{Park19-specaug} policy, a frequency-divided CNN extractor, a layer-wise transfer learning approach, and a pseudo feature pre-training method to bridge the gap between whispered and normal speech.
	We achieve an overall relative reduction of 19.8\% in PER and 44.4\% in CER on a relatively small whispered TIMIT corpus \cite{Lim10-wtimit}, which is already a very narrow gap from the performance for normal speech.

\section{Proposed Methods}
\label{sec:methods}
This work is based on the CTC model~\cite{Graves14-ctcasr} for E2E ASR consisting of a deep CNN feature extractor \cite{Hori17-jointctc} and a multi-layer bidirectional LSTM.
The model takes a sequence of acoustic features $\boldsymbol{x}=(x_1,\dots,x_T)$ with length $T$ for the input utterance.
The sequence is first encoded by the CNN extractor performing downsampling and further by the BLSTMs to obtain a sequence of hidden states.
This sequence is then linearly transformed into $\boldsymbol{y}=(y_1,\dots,y_{T'})$, where each $y_t$ represents a probability distribution over all possible output symbols at each time index, and $T' \leq T$.
The ASR model is trained to minimize the CTC loss function \cite{Graves14-ctcasr, Graves06-ctc}.

\subsection{Analysis for Frequency Importance}
    \label{subsec:observe}
    
    It has been well known that normal speech characteristics are reasonably preserved in whispers for higher frequencies while seriously lost in lower frequencies, as shown in an example in Fig. \ref{fig:feats} \cite{Jovicic08-analcons, Lim10-wtimit, Ito05-analwhsp, Ghaffarzadegan16-pswhsp,  Grozdic17-invfilt, Morris02-recons}.
    We suspect the higher frequencies are more critical to E2E ASR for whispered speech, although both high and low frequencies play essential roles for normal speech.
	We first analyze this assumption here.

	Two E2E ASR pre-trained with normal and whispered speech, respectively, are used for the experiment below.
	We define a learnable weight vector $\mathbf{w}=[w_0~w_1~\dots~w_{\nu-1}]^{\top}$ for all the Mel-frequency bins, where $\nu$ is the total number of the frequency bins of the considered Mel-spectrogram.
	This vector $\mathbf{w}$ is first transformed into a probability distribution by softmax, $\hat{\mathbf{w}}=[\hat{w}_0~\hat{w}_1~\dots~\hat{w}_{\nu-1}]^{\top}=\mathrm{softmax}(\mathbf{w})$, then used to weight the respective Mel-filterbank features, 
	\begin{equation}
		x_{t,f}'=x_{t,f}\cdot \exp(-\hat{w}_f/r),
		\label{eqlearn}
	\end{equation}
	where $x_{t,f}$ is the feature for the $f^{\mathrm{th}}$ Mel-frequency bin at time $t$, $x_{t,f}'$ is the weighted value, and $r$ is a positive scaling factor. 
	The weighted features are then fed to a pre-trained E2E ASR with frozen weights to learn the distribution $\hat{\mathbf{w}}$ for maximizing the CTC loss function, which is supposed to be minimized.
	So those frequency bins suppressed more by higher weights are those more critical for ASR.
	We use stochastic gradient ascent to obtain the learnable weight $\hat{\mathbf{w}}$.
	
	\begin{figure}[t]
	\centering
	\begin{subfigure}[b]{0.49\linewidth}
		\centering
		\includegraphics[width=\linewidth]{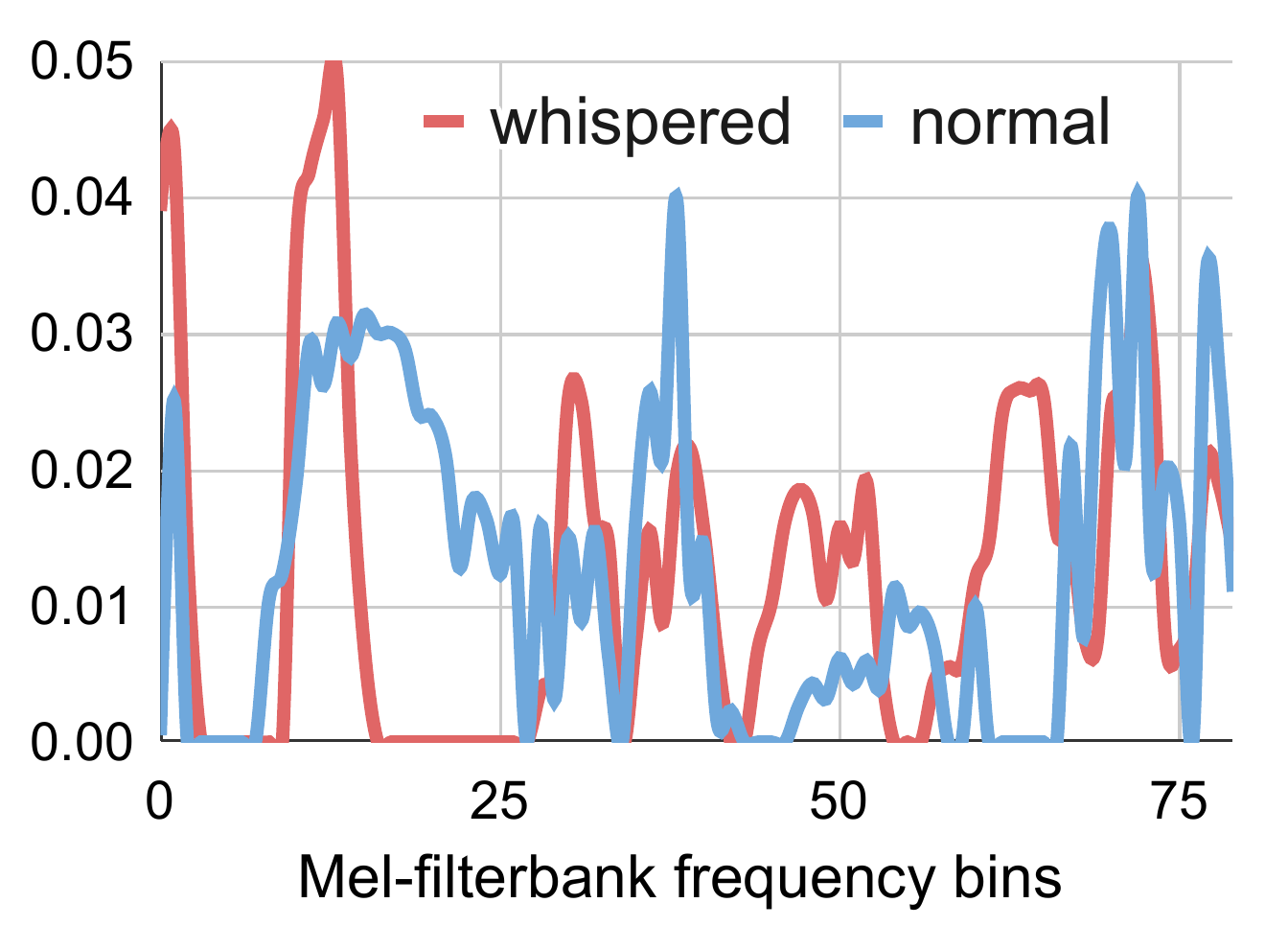}
		\caption{~}
		\label{fig:mask_distrib_freq}
    \end{subfigure}
    \hfill
	\begin{subfigure}[b]{0.49\linewidth}
	    \centering
		\includegraphics[width=\linewidth]{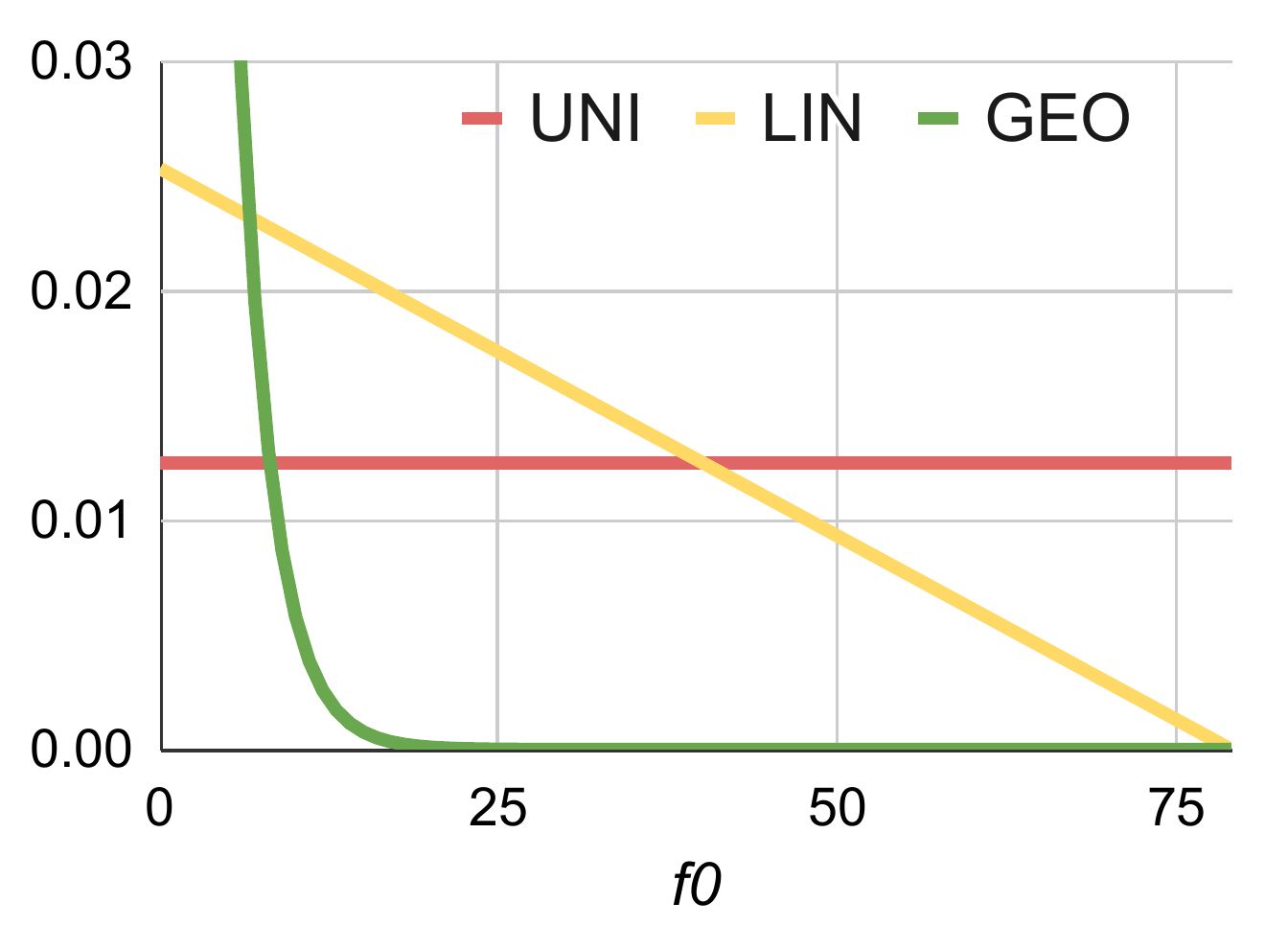}
		\caption{~}
		\label{fig:mask_distrib_f0}
    \end{subfigure}
	\vspace{-6pt}
	\caption{(a) The weight distributions $\hat{\mathbf{w}}$ obtained in the experiment described in Sec. \ref{subsec:observe} for whispered (red) and normal (blue) speech, (b) the uniform (UNI), linearly (LIN), and geometrically (GEO) decreasing distributions for sampling the lower end $f_0$ of the mask in SpecAugment.}
	\label{fig:mask_distrib}
	\vspace{-4pt}
\end{figure}
	With the experimental setup to be described in Sec. \ref{subsec:exp_setup}, the results for the weight distribution $\hat{\mathbf{w}}$ are in Fig. \ref{fig:mask_distrib}(a).
	The learned weights are different for E2E ASR for whispered and normal speech.
	For whispered speech, relatively more emphasis is on higher frequencies, indicating more important information is there.
	In contrast, the weights of normal speech distribute more evenly along the frequency axis.
	Although the analysis may be slightly inaccurate and subjective, this gives us clues for designing different methods proposed below.

\subsection{Frequency-weighted SpecAugment}
	\label{subsec:specaug}
	Frequency masking of SpecAugment \cite{Park19-specaug} has shown useful for data augmentation for E2E ASR models.
	It is summarized as follows.
	A mask size of $\Delta f$ is first sampled from a frequency range $[F_1, F_2]$ uniformly.
	The lower end of the mask, $f_0$, is then sampled uniformly from $[0,\nu-\Delta f)$, where $\nu$ is the total number of frequency bins of the spectrogram.
	These parameters define the mask $[f_0,f_0+\Delta f)$, in which all frequency bins are set to zero when masked.
	
	With the observation in Fig. \ref{fig:mask_distrib}(a), we try to mask lower frequencies more often for whispered speech.
	This is referred to as \textit{Frequency-weighted SpecAugment}, in which instead of sampling $f_0$ uniformly from $[0, \nu-\Delta f)$, $f_0$ can be sampled from a linearly or geometrically decreasing distribution as shown in Fig. \ref{fig:mask_distrib}(b).
	The probability of lower frequency bins being masked would be higher, or the machine would learn less precise information or rely less on lower frequencies.

        \begin{figure}[t]
			\centering
			\includegraphics[width=\linewidth]{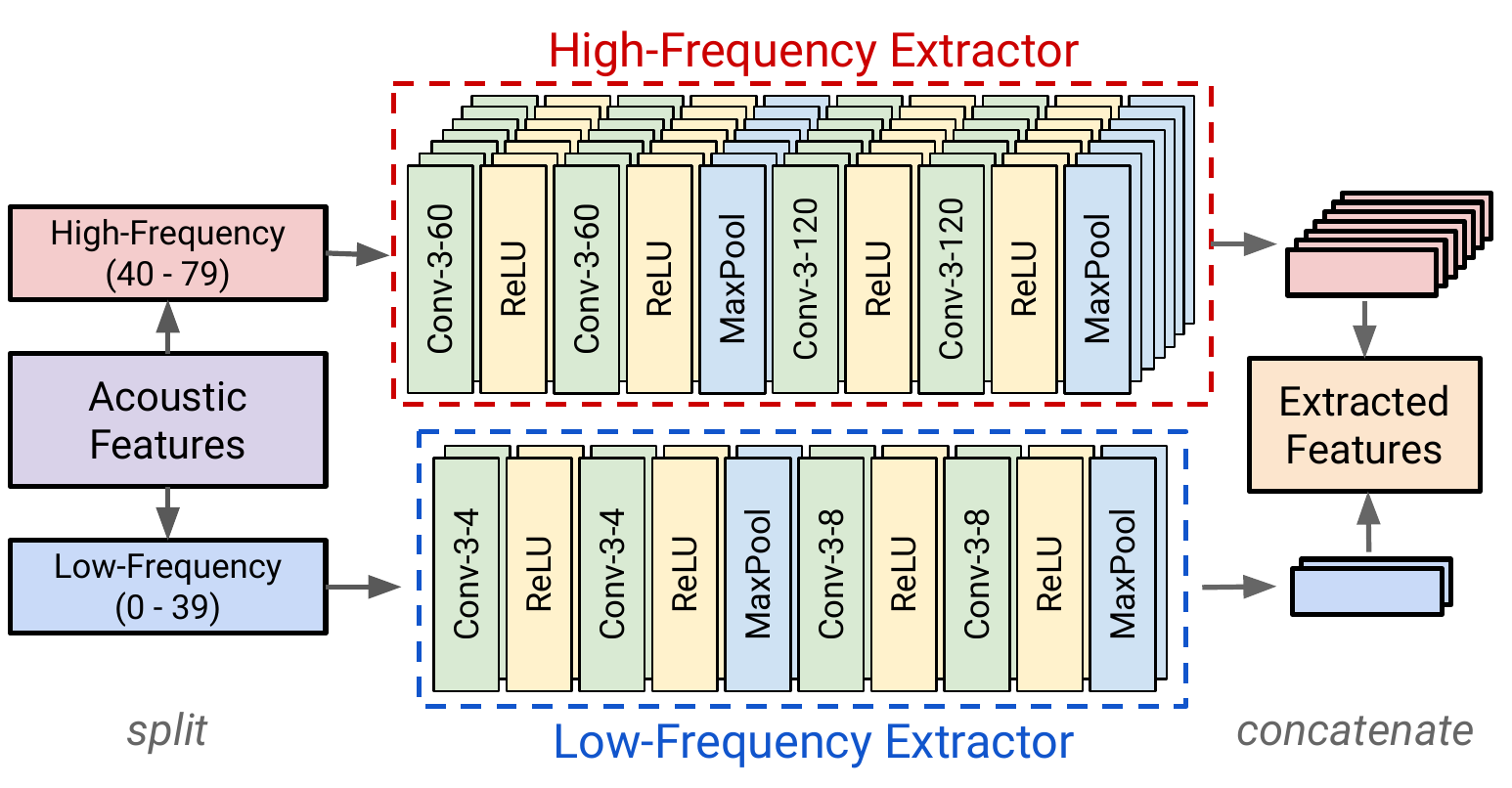}
			\caption{The frequency-divided CNN extractor. \textit{Conv-k-c} denotes 2D convolution with a kernel size of \textit{k} $\times$ \textit{k} and \textit{c} output channels. The low-frequency extractor has fewer convolutional filters to compress the features.}
			\label{fig:freqvgg}
		\end{figure}
\subsection{Frequency-divided CNN Extractor}
	\label{subsec:vgg}
	Since the standard CNN extractor \cite{Hori17-jointctc} used for E2E ASR treats all frequencies equally, here we propose a \textit{Freqency-divided CNN extractor} containing two CNN extractors respectively processing the lower and higher frequency half of the features separately as shown in Fig. \ref{fig:freqvgg}.
	With the same total number of feature parameters as the standard extractor, the low-frequency extractor has fewer filters.
	Therefore, the high-frequency extractor with more filters can capture more cues and offer more information from the preserved structures in high-frequency regions of whispered speech.
	
\subsection{Layer-wise Transfer Learning from Normal Speech}
	\label{subsec:transfer_learning}
	The scarcity of whispered speech data makes training E2E ASR challenging; however, much more normal speech data are available.
	Therefore, we propose to perform transfer learning \cite{Kunze17-transfer} by having an E2E ASR model pre-trained on a large normal speech corpus (Fig. \ref{fig:vc_asr}(a)(I)), and then fine-tuned with a smaller whispered speech corpus (Fig. \ref{fig:vc_asr}(b)).
    However, BLSTMs are prone to overfit~\cite{Nguyen20-improvesq2sq}, fine-tuning the whole model did not work well.
    Since the objective is to transfer between speech types with differences primarily in acoustic characteristics, we propose to fine-tune only the bottom layers closer to the acoustic features, as shown in Fig. \ref{fig:vc_asr}(b).
    This layer-wise transfer learning is similar to but different from that reported for transfer between different languages, in which fine-tuning top layers allow better transfer~\cite{Kunze17-transfer}.

\subsection{Pseudo Whispered Speech for Model Pre-training}
    \label{subsec:vc_method}
    
\begin{figure}[t]
	\centering
	\includegraphics[width=0.95\linewidth]{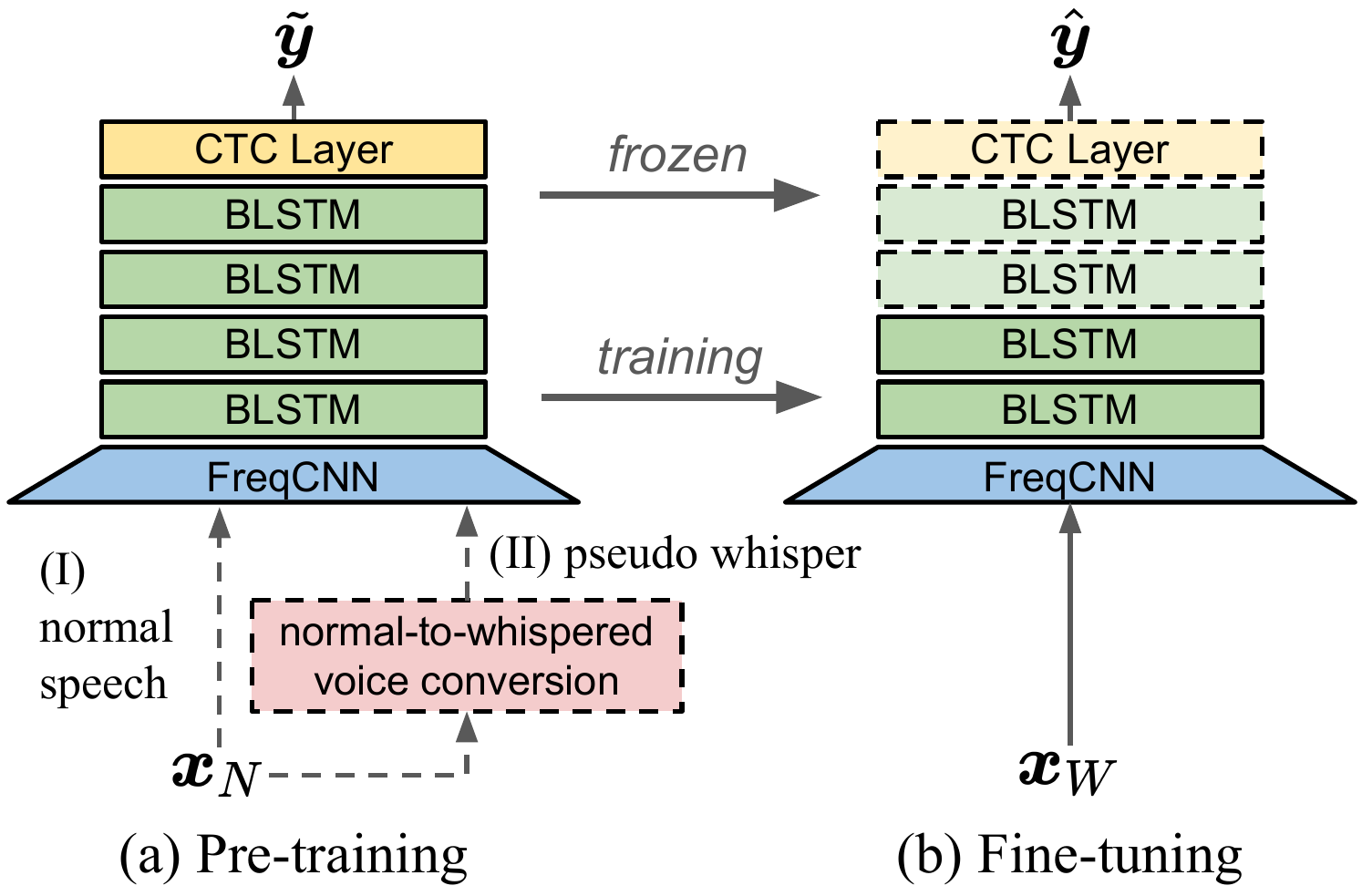}
	\caption{
	The training framework for E2E ASR with layer-wise transfer learning.
	(a) The ASR is first pre-trained with either (I) normal speech $\boldsymbol{x}_N$ or (II) pseudo whispered features generated by a normal-to-whispered voice conversion model.
	(b) Next, the top layers of the ASR are fixed while the lower layers are fine-tuned with real whispered features $\boldsymbol{x}_W$.
	}
	\label{fig:vc_asr}
	\vspace{-4pt}
\end{figure}

    To deal with the scarcity of whispered speech, data-based approaches such as converting normal speech into pseudo whispered features \cite{Ghaffarzadegan16-pswhsp, Grozdic17-invfilt, Ghaffarzadegan17-dnnsmall} or the opposite way \cite{Biadsy19-parrot, Niranjan20-whaletrans} for data augmentation were developed.
    As in Fig. \ref{fig:vc_asr}(a)(II), we first train a voice conversion (VC) model to convert normal speech to whispered acoustic features in a supervised fashion, and then apply this model to a large normal speech corpus to generate pseudo whispered data for ASR pre-training.
    Then fine-tune the bottom layers of the ASR model with a small amount of real whispered speech, as in Fig. \ref{fig:vc_asr}(b).
    We expect a large amount of pseudo whispered data helpful.
    
    Overall, we analyze whispered speech characteristics and exploit them to develop novel approaches to narrow down the performance gap between whispered and normal speech recognition.

\section{Experiments}
\label{sec:exp}

In this section, we purposely organized to step-to-step verify that E2E whispered recognition is feasible and potential.
First, in Sec. \ref{subsec:whsp_exp}, we provided a better corpus partition suitable for evaluating our hypothesis.
Results in Sec. \ref{subsec:specaugexp} showed that it is possible to achieve whispered ASR with limited normal speech and frequency-weighted approaches.
Moreover, layer-wise transfer learning with a small set of whisper significantly improved the performance of whispered ASR as shown in Sec. \ref{subsec:transfer_exp}.
Last, in Sec. \ref{subsec:vc_exp}, to find out the best achievable performance, pseudo-data pre-training and an auxiliary language model was used to show that recognizing whispered speech is achievable and promising.

\subsection{Experimental Setup}
	\label{subsec:exp_setup}
	The following two datasets were used in the experiments:\\
    \noindent\textbf{wTIMIT.} The whispered TIMIT corpus \cite{Lim10-wtimit} consisted of parallel whispered and normal speech data each around 26 hours, including 48 speakers whispering and speaking 450 phonetically balanced sentences chosen from TIMIT \cite{Garofolo93-timit}. 
	This corpus was originally partitioned into the train/test sets randomly.
	However, the existence of many overlapping utterances with the same sentences spoken by different speakers made it challenging to estimate the actual recognition accuracy, as will be shown later in Table \ref{tab:whisper}.
	We thus re-partitioned the dataset into train/dev/test sets; each containing 400/25/25 sentences split from the 450 sentences.
    Since there is still speaker overlap between the three sets, we conducted a preliminary experiment with the corpus partitioned by both speakers and sentences.
    Results showed that partitioning by speakers only degraded the performance by relatively 10\% compared to the case with speaker overlap because pitches in whispered speech are mostly gone.
    Also, partitioning by speakers reduces available data, making training more difficult.
    Therefore, we believe considering the speaker overlap problem is unnecessary.
		
	\noindent\textbf{LibriSpeech.} The LibriSpeech corpus \cite{Panayotov15-libri} included roughly 960 hours of speech.
	The 460-hr set was for normal speech pre-training and 960-hr for pseudo whisper pre-training.
	
	For comparing with HMM-based ASR, a DNN-HMM hybrid system \cite{Hinton12-dnnhmm} baseline was constructed using the TIMIT recipe \textit{nnet2} from the Kaldi toolkit \cite{Povey11-kaldi}.
	13-dimensional MFCC features with delta and delta-delta were used as the recipe did.
	For E2E ASR, 80-dimensional log Mel-filterbank features with delta and normalization were used. 
	Two E2E ASR models were used.
	The standard model used a 4-layered BLSTM of 512 units per direction and a CNN feature extractor \cite{Hori17-jointctc}.
	Another light model with a 3-layered bidirectional GRU with 128 units per direction was used for limited training with wTIMIT (Sec. \ref{subsec:whsp_exp} and \ref{subsec:specaugexp}) to prevent overfitting.

\subsection{E2E \& Hybrid for Whispered Trained on Whispered}
	\label{subsec:whsp_exp}
	        \begin{table}[t]
			\caption{PERs(\%) on whispered for hybrid and E2E ASR trained on wTIMIT whispered with two corpus partitions.}
			\label{tab:whisper}
			\centering
			\begin{tabular}{l|l|cc}
				\toprule
				\multirow{2}{*}{\textbf{Corpus Partition}} & \multirow{2}{*}{\textbf{ASR Model}} & \multicolumn{2}{c}{\textbf{Whispered}} \\
				& & \textbf{Dev} & \textbf{Test} \\
				\hline
				\multirow{2}{*}{(I) Original} & (a) HMM-Hybrid & 35.0 & 35.5\Tstrut\\
				& (b) E2E & 13.0 & 13.7 \\
				\hline
				\multirow{2}{*}{(II) Ours (400/25/25)} & (c) HMM-Hybrid & 39.6 & 38.7\Tstrut \\
				& (d) E2E & 35.6 & 35.9 \\
				\bottomrule
			\end{tabular}
			\vspace{-3pt}
		\end{table}
	We first compared the HMM-based hybrid model with E2E ASR without adopting any proposed approach, assuming only the whispered part of wTIMIT was available for training.
	This paper's HMM-Hybrid baselines are served as references to show how E2E models behave on this new task.
	The phoneme level (total 39 phonemes) annotation was used to train both models from scratch since wTIMIT is very small, and using character or word level was unsuitable.
	The light E2E ASR model used to produce Fig. \ref{fig:mask_distrib}(a) was used here.
	The phoneme error rates (PER) are in Table \ref{tab:whisper}.
	
	Section (I) of Table \ref{tab:whisper} is for the original corpus partition provided by wTIMIT, in which the E2E ASR offered a low error rate (row (b)) as a result of the overlapping utterances between the train/test sets.
	Therefore, all experiments below were based on our partition, as mentioned in Sec. \ref{subsec:exp_setup}, with results in Section (II).
	Here, the E2E model was slightly better than the hybrid (rows (d) v.s. (c)), even with only 26 hours of training data, for which hybrid typically outperformed E2E.
	These results indicated that E2E ASR was a proper choice for whispered speech if the data set was not too small.

\subsection{Proposed Frequency-weighted Approaches with Limited Normal Speech Training}
	\label{subsec:specaugexp}
	
	We considered the case when only limited normal speech (26 hours of normal speech from wTIMIT) was available for training E2E ASR to verify if the proposed frequency-weighted approaches were useful for whispered speech regardless of the training data.
	The light model same as that used in Sec. \ref{subsec:whsp_exp} was used.
	The results are listed in Table~\ref{tab:specaug}.
	
	\begin{table}[t]
	\caption{PERs(\%) on wTIMIT with only a small normal speech set for training. 
	Section (I) for baselines, Section (II) with the proposed frequency-weighted SpecAugment (FreqSpecAug) with a uniform (UNI), linearly (LIN), and geometrically (GEO) decreasing distribution, Section (III) with frequency-divided CNN extractor (FreqCNN) applied.}
	\label{tab:specaug}
	\centering
	\begin{tabular}{ll|cc|cc}
		\toprule
		\multicolumn{2}{l|}{\multirow{2}{*}{\textbf{Method}}} & \multicolumn{2}{c|}{\textbf{(A) Normal}~~~~~} & \multicolumn{2}{c}{\textbf{(B) Whispered}} \\
		& & \textbf{Dev} & \textbf{Test} & \textbf{Dev} & \textbf{Test} \\
		\hline
		\multicolumn{6}{l}{\textbf{(I) Baselines\Tstrut}} \\
		\hline
		~ & (a) HMM-Hybrid & 34.5 & 33.5 & 55.8 & 54.6\Tstrut \\
		~ & (b) E2E & \textbf{29.9} & \textbf{29.7} & 60.7 & 59.5 \\
		\hline
		\multicolumn{6}{l}{\textbf{(II) E2E + FreqSpecAug\Tstrut}} \\
		\hline
		~ & (c) UNI~\cite{Park19-specaug} & 31.0 & 31.1 & 54.8 & 53.9\Tstrut \\
		~ & (d) LIN & 30.6 & 30.6 & 52.9 & 51.8 \\
		~ & (e) GEO & 33.1 & 32.8 & 49.2 & 48.3 \\
		\hline
		\multicolumn{6}{l}{\textbf{(III) E2E + FreqCNN + FreqSpecAug\Tstrut}} \\
		\hline
		~ & (f) UNI & 33.5 & 32.8 & 52.0 & 51.3\Tstrut \\
		~ & (g) GEO & 35.5 & 35.1 & \textbf{48.3} & \textbf{47.7} \\
		\bottomrule
	\end{tabular}
	\vspace{-8pt}
\end{table}

	\noindent\textbf{Trained with Limited Normal Speech}
	Section (I) of Table \ref{tab:specaug} is for the baselines with the zero whispered speech resource scenario without adopting any approach proposed here.
	The model performance degraded seriously for whispered speech (columns (B) v.s. (A)), and E2E performed better on normal speech yet worse on whispered (rows (b) v.s. (a)).
	These results aligned with the mismatch between whispered and normal and the assumption that E2E ASR was prone to overfit its training data \cite{Nguyen20-improvesq2sq}.
		
	\noindent\textbf{Frequency-weighted SpecAugment}
	To find out the extra robustness achievable by the proposed frequency-weighted SpecAugment, we let the lower end $f_0$ of the mask in SpecAugment to be sampled from a uniform (UNI), linearly (LIN), or geometrically (GEO) decreasing distribution as described in Sec. \ref{subsec:specaug}.
	The results are in rows (c)(d)(e) of Section (II) in Table \ref{tab:specaug}.
	With GEO proposed, a relative 18.8\% PER reduction for the baseline E2E (rows (e) v.s. (b)) and a 10.4\% relative improvement compared to the original SpecAugment or UNI (rows (e) v.s. (c)) was achieved.
	These results implied with the lower frequencies emphasized in SpecAugment or letting E2E ASR learn less from lower frequencies, the performance on whispered speech was improved.
	Thus, the model distilled more details from higher frequencies where whispered is more similar to normal.
	
	\noindent\textbf{Frequency-divided CNN Extractor}
	In Section (III) of Table \ref{tab:specaug} we added the frequency-divided CNN extractor as mentioned in Sec. \ref{subsec:vgg} onto the models in Section (II).
	Results in rows (f)(g) showed that the proposed extractor made further PER reduction on whispered speech (rows (g) v.s. (e) and (f) v.s. (c)).
	Thereby, verified that extracting less low-frequency information with fewer filters while more high-frequency information with more filters did help.
	However, this approach inevitably degraded the accuracy of normal speech.
	The overall relative improvement achieved by the frequency-weighted SpecAugment plus frequency-divided CNN extractor was 19.8\% (rows (g) v.s. (b)), which was the setting for the experiments below.

\subsection{Training with Extra Normal Speech Data}
	\label{subsec:transfer_exp}
	Here, we tried to reduce the performance gap between whispered and normal speech recognition using an additional large normal speech corpus (LibriSpeech).
	The models were trained on grapheme level (characters without lexicon) for real-world applications, and following previous works \cite{Collobert16-wav2letter, Liptchinsky17-letterbased, Likhomanenko19-whoneedswords}, with frequency-weighted SpecAugment and frequency-divided CNN extractor applied.
		
\begin{figure}[t]
			\centering
			\includegraphics[width=\linewidth]{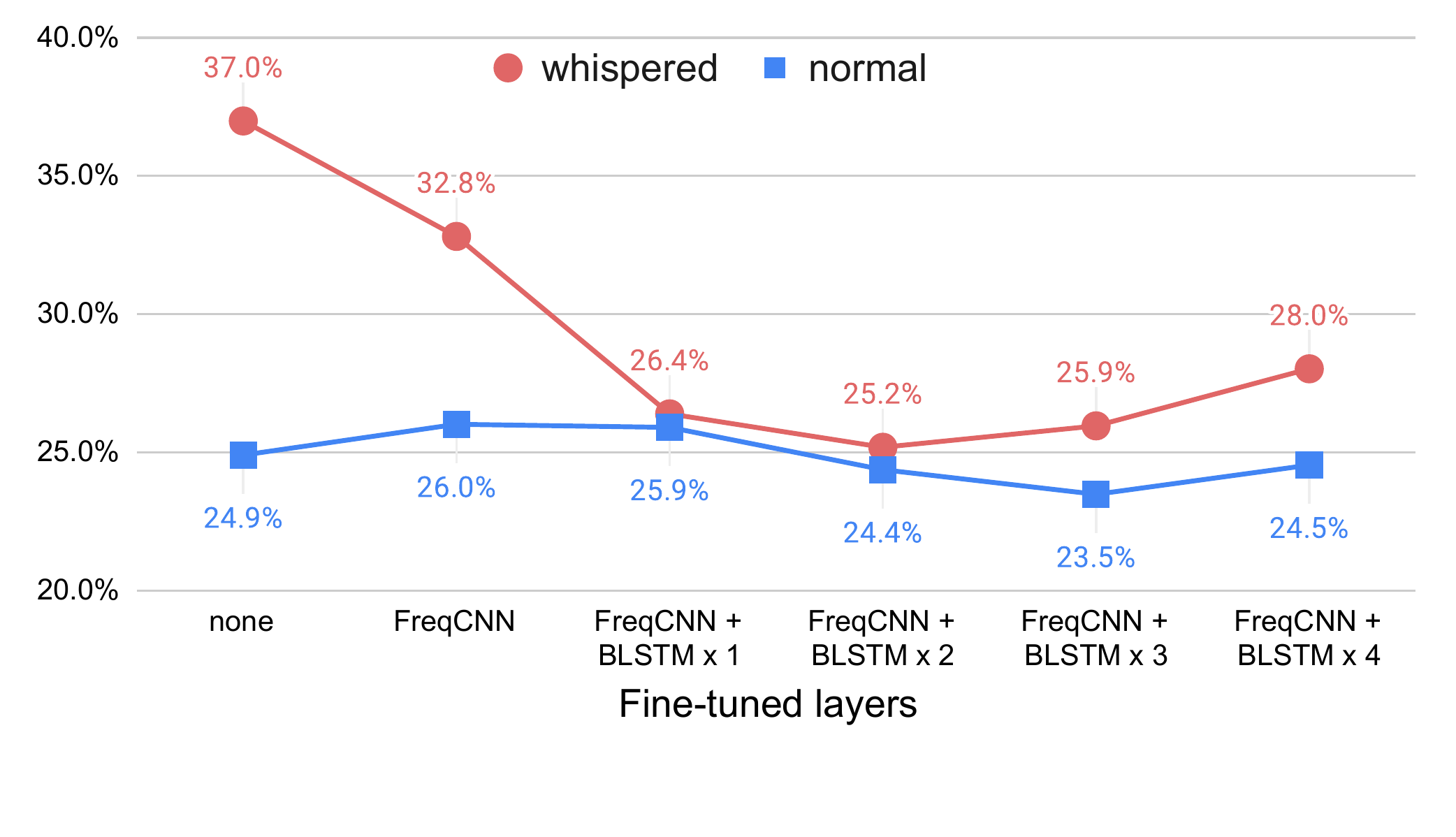}
			\vspace{-37pt}
			\caption{CERs for whispered and normal on wTIMIT when fine-tuned from the bottom in the layer-wise transfer learning.}
			\label{fig:transfer}
			\vspace{-4pt}
		\end{figure}

    \noindent\textbf{Layer-wise Transfer Learning}
    Here we studied the layer-wise transfer learning mentioned in Section \ref{subsec:transfer_learning}.
    We used the 460-hr LibriSpeech normal speech data to pre-train an E2E ASR model and then fine-tuned it with the whispered speech in wTIMIT.
    Instead of fine-tuning the whole model, only several bottom layers were fine-tuned.
    Fig. \ref{fig:transfer} depicts the results for whispered and normal speech from left to right when fine-tuning was performed on a different number of bottom layers, starting with no fine-tuning.
    The fine-tuning procedure used stochastic gradient descent with a fixed learning rate.
        
    In Fig. \ref{fig:transfer}, the character error rate (CER) for whispered speech was improved from 37.0\% to 25.2\% when the frequency-divided CNN extractor and the first two BLSTM layers were fine-tuned simultaneously, which was a 31.9\% error rate reduction relative to the pre-trained model.
    Fine-tuning the 3rd BLSTM layer or further did not boost the performance.
    This is probably because layers close to the output were more related to characters and language modeling \cite{Kunze17-transfer}, and fine-tuning too many parameters affected the model's ability and further overfitted it on the small wTIMIT corpus.
	We also tried to fine-tune the output layer; however, that slightly damaged the performance, possibly because we recognized the same language.
	The best result of 25.2\% here was only 1.7\% absolutely higher than the best performance on normal speech when fine-tuning an extra layer.
	These results verified that the BLSTMs played essential roles in encoding acoustic features, and thus fine-tuning part of the bottom layers of a pre-trained model was helpful.
		
	\noindent\textbf{Different Methods for Training with Both Speech Types}
	        \begin{table}[t]
			\caption{CERs(\%) on wTIMIT when an additional normal corpus is available.
			wTM-w and wTM-n (rows(a)(b)) denote whispered and normal data from wTIMIT, respectively.
			Libri (row(c)) denotes the LibriSpeech 460-hr set as the additional data.
			Imbalanced learning (row(e)) is the previously used method \cite{Ashihara19-whspdetec}.
			Layer-wise TL (row(f)) denotes the layer-wise transfer learning proposed here.}
			\label{tab:ctc_cer}
			\centering
			\begin{tabular}{@{\hskip 1pt}l@{\hskip 2pt}|l@{\hskip 2pt}|@{\hskip 1pt}c@{\hspace{1pt}}|c@{\hskip 6pt}c|c@{\hskip 6pt}c@{\hskip 1pt}}
				\toprule
				\multicolumn{2}{l|}{\multirow{2}{*}{\textbf{Method}}} & \multicolumn{1}{@{\hskip 2pt}c@{\hskip 2pt}|}{\multirow{2}{*}{\shortstack{\textbf{Training}\\\textbf{Data}}}} & \multicolumn{2}{@{\hskip 2pt}c@{\hskip 2pt}|}{\textbf{Normal}} & \multicolumn{2}{@{\hskip 2pt}c@{\hskip 2pt}}{\textbf{Whispered}} \\
				\cline{4-7}
				\multicolumn{2}{l|}{}&& \textbf{Dev} & \textbf{Test} & \textbf{Dev} & \textbf{Test}\Tstrut \\
				\hline
				(a) & \multicolumn{1}{@{\hskip 2pt}l|}{\multirow{3}{*}{\shortstack{E2E baselines\\(single dataset)}}} & wTM-w & 54.4 & 53.1 & 48.0 & 46.1\Tstrut \\ \cline{1-1}\cline{3-7}
				(b) &\multicolumn{1}{@{\hskip 2pt}l|}{}& wTM-n & 41.9 & 40.5 & 54.8 & 53.3\Tstrut \\ \cline{1-1}\cline{3-7}
				(c) &\multicolumn{1}{@{\hskip 2pt}l|}{}& Libri & 26.4 & 24.9 & 37.8 & 37.0\Tstrut \\ \cline{1-1}\cline{3-7}
				\hline
				\multicolumn{2}{@{\hskip 1pt}l@{\hskip 2pt}|}{(d) Random Sampling} & \multicolumn{1}{@{\hskip 2pt}c@{\hskip 2pt}|}{\multirow{3}{*}{\shortstack{wTM-wn\\+\\Libri}}} & 28.7 & 28.1 & 34.0 & 32.9\Tstrut \\ \cline{1-2}\cline{4-7}
				\multicolumn{2}{@{\hskip 1pt}l@{\hskip 2pt}|}{(e) Imbalanced learning} && 43.6 & 41.7 & 47.4 & 45.2\Tstrut \\ \cline{1-2}\cline{4-7}
				\multicolumn{2}{@{\hskip 1pt}l@{\hskip 2pt}|}{(f)~ Layer-wise TL} && \textbf{24.4} & \textbf{23.5} & \textbf{26.5} & \textbf{25.2}\Tstrut \\
				\bottomrule
			\end{tabular}
			\vspace{-8pt}
		\end{table}

	We wish to explore different methods using both whispered and normal speech to train the E2E ASR for whispered speech.
	We first set three baseline models trained solely on the wTIMIT whispered set (wTM-w), the wTIMIT normal set (wTM-n), and the LibriSpeech corpus separately, respectively in rows (a)(b)(c) of Table \ref{tab:ctc_cer}.
	We then used all the three sets of whispered and normal speech jointly to train the E2E ASR in rows (d)(e)(f) in the 2nd half of Table \ref{tab:ctc_cer}.
	This included directly sampling them randomly regardless of the size of the corpus (row (d)), oversampling whispered speech to the same size as normal speech (referred to as imbalanced learning, previously used for whisper detection \cite{Ashihara19-whspdetec}) (row (e)), and the layer-wise transfer learning in Fig. \ref{fig:transfer} (row (f)).
	
	First of all, the baseline models using wTIMIT performed poorly compared to using LibriSpeech (rows (a)(b) v.s. (c)), confirming that E2E models required a large amount of training data to work well \cite{Han19-sotaasr}.
	Next, mixing all whispered and normal speech data, the performance improved slightly on the whispered set while degraded on the normal set compared to the model using only normal speech (rows (d) v.s. (c)).
	Though with a relatively small amount of whispered speech (only about 5\%), the E2E ASR model still learned to recognize whispered speech.
	Moreover, the imbalanced learning damaged the E2E ASR severely (row (e)), perhaps due to the low diversity of the sentences in wTIMIT; the system thus failed to model characters and words.
		
	In contrast, for the layer-wise transfer learning method (row (f)), we divided the training phase into two, pre-training with normal speech and fine-tuning with whispered speech.
	This method outperformed all other methods. 
	Based on the model well-initialized with a sizeable normal set, fine-tuning a part of its layers adapted it to whispered speech while preserving its original capability to recognize the vocabulary's various words.
	In other words, the layer-wise transfer learning proposed here enables us to bridge the gap between recognizing normal and whispered speech.
	Therefore, we can use any E2E model pre-trained on normal speech without collecting a vast amount of whispered speech.

\subsection{Training with Pseudo Whispered Features}
    \label{subsec:vc_exp}
    \begin{figure}[t]
	\centering
	\begin{subfigure}[b]{\linewidth}
	    \centering
		\includegraphics[width=0.8\linewidth, height=32pt]{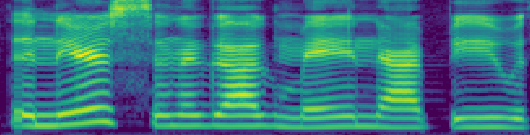}
		\caption{normal Mel-spectrogram}
	    \label{fig:vc_feat_norm}
    \end{subfigure}
    \\
    \vspace{3pt}
    \begin{subfigure}[b]{\linewidth}
	    \centering
		\includegraphics[width=0.8\linewidth, height=32pt]{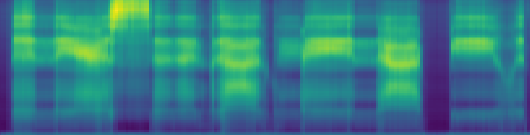}
		\caption{pseudo whispered Mel-spectrogram}
	    \label{fig:vc_feat_out}
    \end{subfigure}
    \\
    \vspace{3pt}
    \begin{subfigure}[b]{\linewidth}
	    \centering
		\includegraphics[width=0.8\linewidth, height=32pt]{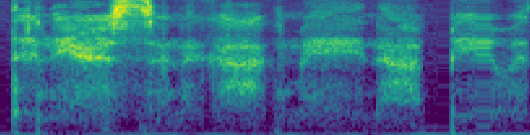}
		\caption{ground-truth whispered Mel-spectrogram}
	    \label{fig:vc_feat_whsp}
    \end{subfigure}
	\vspace{-19pt}
	\caption{
	The result of converting (a) normal speech in wTIMIT to (b) pseudo whispered speech with a DNN-based VC model, where the ground-truth whispered speech Mel-spectrogram is shown in (c).
	}
	\label{fig:vc_feat}
	\vspace{-7pt}
\end{figure}
    This section further evaluates pre-training effectiveness with pseudo whispered speech followed by layer-wise transfer learning as mentioned in Section \ref{subsec:vc_method}.
    
    \noindent\textbf{VC Model}
    First, we built a 4-layered deep neural network VC model \cite{Cotescu19-whspvc} and trained with clean paired whispered-normal utterances aligned with FastDTW \cite{Salvador07-fastdtw}.
    Data were chosen from 25 manually selected speakers in wTIMIT because the recording quality across speakers varied significantly.
    
    An example of the conversion results of the VC model is shown in Fig. \ref{fig:vc_feat}.
    Given the input normal utterance in Fig. \ref{fig:vc_feat}(a), comparing the output pseudo whispered features in Fig. \ref{fig:vc_feat}(b) with the ground-truth whispered features in Fig. \ref{fig:vc_feat}(c), the major features in the ground-truth whispered spectrogram could be generated from the VC model.
    However, some differences between the two are visible.
    This model was applied to the pre-training of E2E ASR below.
    
    \noindent\textbf{ASR Pre-training with Pseudo Whispered Speech}
    \begin{table}[t]
	\caption{
	CERs(\%) on wTIMIT for pre-training with (I) frequency-weighted SpecAugment (960 hours), (II) pseudo whisper (960 hours), and (III) both.
	After pre-training, the models are fine-tuned with layer-wise transfer learning (rows (b)(d)(f)).
	Column (B) with RNN-LM applied in addition.
	}
	\label{tab:vc_cer}
	\centering
	\begin{tabular}{l|cc|cc}
		\toprule
		\multirow{2}{*}{\textbf{Method}} & \multicolumn{2}{c|}{\textbf{(A) w/o LM}} & \multicolumn{2}{c}{~\textbf{(B) w/ LM}~} \\
		& \textbf{Dev} & \textbf{Test} & \textbf{Dev} & \textbf{Test} \\
		\hline
		\multicolumn{5}{l}{\textbf{(I) FreqSpecAug\Tstrut}} \\
		\hline
		~~(a) Pre-trained & 37.9 & 36.9 & 35.7 & 34.2 \Tstrut \\
		~~(b) Layer-wise TL & 25.0 & 23.8 & 21.6 & 19.8 \Tstrut \\
		\hline
		\multicolumn{5}{l}{\textbf{(II) Pseudo Whisper\Tstrut}} \\
		\hline
		~~(c) Pre-trained & 56.2 & 55.6 & 53.6 & 53.3 \Tstrut \\
		~~(d) Layer-wise TL & 24.4 & 23.4 & 21.0 & 19.9 \Tstrut \\
		\hline
		\multicolumn{5}{l}{\textbf{(III) Mixed FreqSpecAug \& Pseudo Whisper\Tstrut}} \\
		\hline
		~~(e) Pre-trained & 37.1 & 35.7 & 34.6 & 33.2 \Tstrut \\
		~~(f) Layer-wise TL & \textbf{24.0} & \textbf{22.5} & \textbf{20.8} & \textbf{19.0} \Tstrut \\
		\bottomrule
	\end{tabular}
	\vspace{-6pt}
\end{table}

    Here, we examined the proposed pseudo whispered speech pre-training method.
    To find out the best achievable performance, we used all 960-hour normal data in LibriSpeech for pre-training.
    We then performed CTC beam decoding \cite{Graves14-ctcasr} rescored with an RNN-based language model \cite{Sundermeyer15-RNNLM} also trained with LibriSpeech.
    The results are listed in Table \ref{tab:vc_cer}.
    
    First in Section (I) of Table \ref{tab:vc_cer} for pre-training with frequency-weighted SpecAugment on the 960 hours of normal speech, by increasing the data for pre-training to 960 hours, the ASR was improved slightly for whispered speech (36.9\% in column (A), row (a) of Table \ref{tab:vc_cer} v.s. 37.0\% in row (c) of Table \ref{tab:ctc_cer}).
    In contrast, without frequency-weighted SpecAugment while only performing the pseudo whisper pre-training, Section (II) of Table \ref{tab:vc_cer} showed terrible error rates were obtained compared to SpecAugment (rows (c) v.s (a)).
    The VC model probably caused this phenomenon since it was challenging to generate features the same as whispered.
    
    Next, with layer-wise transfer learning applied, the ASR performance was significantly improved in either Section (I) or (II) (rows (b) v.s. (a) and (d) v.s. (c)).
    Moreover, the pseudo whisper pre-trained ASR surpassed frequency-weighted SpecAugment (rows (d) v.s. (b)), verifying the model adapted better by pre-training with pseudo data.
    In Section (III), we further mixed the two types of data (960 hours of normal speech with FreSpecAug plus another 960 hours of pseudo whisper, a total of 1920 hours generated from the 960 hours of normal speech only) for pre-training.
    Even lower CERs on whispered speech were obtained (Sections (III) v.s. (II)(I)).
    So FreqSpecAug in Section (I) and Pseudo Whisper in Section (II) were actually complementary.
    With RNN-LM further applied in column (B) of Table \ref{tab:vc_cer}, the best CER was lowered to 19.0\%, which was a relative reduction of 44.4\% compared to the SpecAugment baseline (rows (f) v.s. (a)).
    Overall, we showed that pre-training E2E ASR with pseudo whisper and decode with an additional RNN-LM could offer good whispered speech performance.

\section{Conclusion}
This paper is the first study exploring the possibility of E2E recognition for whispered speech. We propose a frequency-weighted SpecAugment approach and a frequency-divided CNN extractor to boost the recognition performance.
With the aid of a larger normal speech corpus, a pseudo whisper pre-training method, and a layer-wise transfer learning approach plus RNN-LM assisted beam decoding, we further show that the performance gap between whispered and normal speech recognition can be reduced to very narrow even with a minimal data set for whispered speech.
Furthermore, we believe that this work can serve as a comprehensive reference for future work in this and other related areas.

\bibliographystyle{IEEEtran}

\bibliography{mybib}

\begin{thebibliography}{10}
\providecommand{\url}[1]{#1}
\csname url@samestyle\endcsname
\providecommand{\newblock}{\relax}
\providecommand{\bibinfo}[2]{#2}
\providecommand{\BIBentrySTDinterwordspacing}{\spaceskip=0pt\relax}
\providecommand{\BIBentryALTinterwordstretchfactor}{4}
\providecommand{\BIBentryALTinterwordspacing}{\spaceskip=\fontdimen2\font plus
\BIBentryALTinterwordstretchfactor\fontdimen3\font minus
  \fontdimen4\font\relax}
\providecommand{\BIBforeignlanguage}[2]{{%
\expandafter\ifx\csname l@#1\endcsname\relax
\typeout{** WARNING: IEEEtran.bst: No hyphenation pattern has been}%
\typeout{** loaded for the language `#1'. Using the pattern for}%
\typeout{** the default language instead.}%
\else
\language=\csname l@#1\endcsname
\fi
#2}}
\providecommand{\BIBdecl}{\relax}
\BIBdecl

\bibitem{Jovicic08-analcons}
S.~T. {Jovi{\v{c}}i{\'c}} and Z.~{{\v{S}}ari{\'c}}, ``Acoustic analysis of
  consonants in whispered speech,'' \emph{Journal of Voice}, vol.~22, 2008.

\bibitem{Lim10-wtimit}
B.~P. Lim, ``Computational differences between whispered and non-whispered
  speech,'' Ph.D. dissertation, University of Illinois at Urbana Champaign,
  2010.

\bibitem{Lee14-iwhisp}
P.~X. {Lee}, D.~{Wee}, H.~S.~Y. {Toh}, B.~P. {Lim}, N.~{Chen}, and B.~{Ma}, ``A
  whispered mandarin corpus for speech technology applications,'' in
  \emph{INTERSPEECH}, 2014.

\bibitem{Ito05-analwhsp}
T.~Ito, K.~Takeda, and F.~Itakura, ``Analysis and recognition of whispered
  speech,'' \emph{Speech Communication}, vol.~45, 2005.

\bibitem{Ghaffarzadegan16-pswhsp}
S.~{Ghaffarzadegan}, H.~{Bo{\v{r}}il}, and J.~H.~L. {Hansen}, ``Generative
  modeling of pseudo-whisper for robust whispered speech recognition,''
  \emph{IEEE/ACM Transactions on Audio, Speech, and Language Processing},
  vol.~24, 2016.

\bibitem{Grozdic17-invfilt}
{\Dbar}.~T. {Grozdi{\'c}} and S.~T. {Jovi{\v{c}}i{\'c}}, ``Whispered speech
  recognition using deep denoising autoencoder and inverse filtering,''
  \emph{IEEE/ACM Transactions on Audio, Speech, and Language Processing},
  vol.~25, 2017.

\bibitem{Morris02-recons}
R.~W. {Morris} and M.~A. {Clements}, ``Reconstruction of speech from
  whispers,'' \emph{Medical Engineering \& Physics}, vol.~24, 2002.

\bibitem{Jou05-articulatory}
{Szu-Chen Jou}, T.~{Schultz}, and A.~{Waibel}, ``Whispery speech recognition
  using adapted articulatory features,'' in \emph{ICASSP}, 2005.

\bibitem{Mathur12-param}
A.~{Mathur}, S.~{Reddy}, and R.~{Hegde}, ``Significance of parametric spectral
  ratio methods in detection and recognition of whispered speech,'' in
  \emph{EURASIP Journal on Advances in Signal Processing}, 2012.

\bibitem{Ghaffarzadegan17-dnnsmall}
S.~{Ghaffarzadegan}, H.~{Bo{\v{r}}il}, and J.~H.~L. {Hansen}, ``Deep neural
  network training for whispered speech recognition using small databases and
  generative model sampling,'' \emph{J Speech Technol}, vol.~20, 2017.

\bibitem{Yang12-murmur}
C.~{Yang}, G.~{Brown}, L.~{Lu}, J.~{Yamagishi}, and S.~{King}, ``Noise-robust
  whispered speech recognition using a non-audible-murmur microphone with vts
  compensation,'' in \emph{ISCSLP}, 2012.

\bibitem{Srinivasan19-articulatory}
G.~{Srinivasan}, A.~{Illa}, and P.~K. {Ghosh}, ``A study on robustness of
  articulatory features for automatic speech recognition of neutral and
  whispered speech,'' in \emph{ICASSP}, 2019.

\bibitem{Cao16-acticulatory}
B.~{Cao}, M.~{Kim}, T.~{Mau}, and J.~{Wang}, ``Recognizing whispered speech
  produced by an individual with surgically reconstructed larynx using
  articulatory movement data,'' in \emph{SLPAT}, 2016.

\bibitem{Tao14-lipreading}
F.~{Tao} and C.~{Busso}, ``Lipreading approach for isolated digits recognition
  under whisper and neutral speech,'' in \emph{INTERSPEECH}, 2014.

\bibitem{Tran13-audiovis}
T.~{Tran}, S.~{Mariooryad}, and C.~{Busso}, ``Audiovisual corpus to analyze
  whisper speech,'' in \emph{ICASSP}, 2013.

\bibitem{Petridis18-visual}
S.~{Petridis}, J.~{Shen}, D.~{Cetin}, and M.~{Pantic}, ``Visual-only
  recognition of normal, whispered and silent speech,'' in \emph{ICASSP}, 2018.

\bibitem{Rabiner89-hmm}
L.~R. Rabiner, ``A tutorial on hidden {Markov} models and selected applications
  in speech recognition,'' \emph{Proceedings of the IEEE}, vol.~77, 1989.

\bibitem{Graves14-ctcasr}
A.~{Graves} and N.~{Jaitly}, ``Towards end-to-end speech recognition with
  recurrent neural networks,'' in \emph{ICML}, 2014.

\bibitem{Graves12-rnnt}
A.~Graves, ``Sequence transduction with recurrent neural networks,'' in
  \emph{ICML Workshop on Representation Learning}, 2012.

\bibitem{Chan16-las}
W.~{Chan}, N.~{Jaitly}, Q.~{Le}, and O.~{Vinyals}, ``Listen, attend and spell:
  A neural network for large vocabulary conversational speech recognition,'' in
  \emph{ICASSP}, 2016.

\bibitem{Markovic13-whispe}
B.~{Markovi{\'c}}, S.~T. {Jovi{\v{c}}i{\'c}}, J.~{Gali{\'c}}, and {\Dbar}.~T.
  {Grozdi{\'c}}, ``Whispered speech database: design, processing and
  application,'' in \emph{Text, Speech, and Dialogue}, 2013.

\bibitem{Grozdic13-appwhsp}
{\Dbar}.~T. {Grozdi{\'c}}, B.~{Markovi{\'c}}, J.~{Gali{\'c}}, and S.~T.
  {Jovi{\v{c}}i{\'c}}, ``Application of neural networks in whispered speech
  recognition,'' \emph{Telfor Journal}, vol.~5, 2013.

\bibitem{Lim16-transfer}
B.~P. {Lim}, F.~{Wong}, Y.~{Li}, and J.~W. {Bay}, ``Transfer learning with
  bottleneck feature networks for whispered speech recognition,'' in
  \emph{INTERSPEECH}, 2016.

\bibitem{Amodei16-deep-speech-2}
D.~Amodei, S.~Ananthanarayanan, R.~Anubhai, J.~Bai, E.~Battenberg, C.~Case,
  J.~Casper, B.~Catanzaro, Q.~Cheng, G.~Chen \emph{et~al.}, ``Deep speech 2:
  End-to-end speech recognition in english and mandarin,'' in \emph{ICML},
  2016.

\bibitem{Chiu18-sotaasr}
C.~{Chiu}, T.~N. {Sainath}, Y.~{Wu}, R.~{Prabhavalkar}, P.~{Nguyen}, Z.~{Chen},
  A.~{Kannan}, R.~J. {Weiss}, K.~{Rao}, E.~{Gonina}, N.~{Jaitly}, B.~{Li},
  J.~{Chorowski}, and M.~{Bacchiani}, ``State-of-the-art speech recognition
  with sequence-to-sequence models,'' in \emph{ICASSP}, 2018.

\bibitem{Hori17-jointctc}
T.~Hori, S.~Watanabe, Y.~Zhang, and W.~Chan, ``Advances in joint ctc-attention
  based end-to-end speech recognition with a deep cnn encoder and rnn-lm,'' in
  \emph{INTERSPEECH}, 2017.

\bibitem{Zhang17-very-deep-conv}
Y.~Zhang, W.~Chan, and N.~Jaitly, ``Very deep convolutional networks for
  end-to-end speech recognition,'' in \emph{ICASSP}, 2017.

\bibitem{Han19-sotaasr}
K.~J. {Han}, R.~{Prieto}, and T.~{Ma}, ``State-of-the-art speech recognition
  using multi-stream self-attention with dilated 1d convolutions,'' in
  \emph{ASRU}, 2019.

\bibitem{Park19-specaug}
D.~S. Park, W.~Chan, Y.~Zhang, C.-C. Chiu, B.~Zoph, E.~D. Cubuk, and Q.~V. Le,
  ``{SpecAugment: A Simple Data Augmentation Method for Automatic Speech
  Recognition},'' in \emph{INTERSPEECH}, 2019.

\bibitem{Graves06-ctc}
A.~Graves, S.~Fern\'{a}ndez, F.~Gomez, and J.~Schmidhuber, ``Connectionist
  temporal classification: Labelling unsegmented sequence data with recurrent
  neural networks,'' in \emph{ICML}, 2006.

\bibitem{Kunze17-transfer}
J.~Kunze, L.~Kirsch, I.~Kurenkov, A.~Krug, J.~Johannsmeier, and S.~Stober,
  ``Transfer learning for speech recognition on a budget,'' in
  \emph{Proceedings of the 2nd Workshop on Representation Learning for {NLP}},
  2017.

\bibitem{Nguyen20-improvesq2sq}
T.-S. {Nguyen}, S.~{St\"{u}ker}, J.~{Niehues}, and A.~{Waibel}, ``Improving
  sequence-to-sequence speech recognition training with on-the-fly data
  augmentation,'' in \emph{ICASSP}, 2020.

\bibitem{Biadsy19-parrot}
F.~Biadsy, R.~J. Weiss, P.~J. Moreno, D.~Kanvesky, and Y.~Jia, ``{Parrotron: An
  End-to-End Speech-to-Speech Conversion Model and its Applications to
  Hearing-Impaired Speech and Speech Separation},'' in \emph{INTERSPEECH},
  2019.

\bibitem{Niranjan20-whaletrans}
A.~Niranjan, M.~Sharma, S.~B.~C. Gutha, and M.~Shaik, ``Whaletrans: E2e whisper
  to natural speech conversion using modified transformer network,''
  \emph{arXiv preprint arXiv:2004.09347}, 2020.

\bibitem{Garofolo93-timit}
J.~S. {Garofolo}, L.~F. {Lamel}, W.~M. {Fisher}, J.~G. {Fiscus}, D.~S.
  {Pallett}, and N.~L. {Dahlgren}, ``Timit acoustic-phonetic continuous speech
  corpus,'' \emph{Linguistic Data Consortium}, 1992.

\bibitem{Panayotov15-libri}
V.~{Panayotov}, G.~{Chen}, D.~{Povey}, and S.~{Khudanpur}, ``Librispeech: An
  asr corpus based on public domain audio books,'' in \emph{ICASSP}, 2015.

\bibitem{Hinton12-dnnhmm}
G.~{Hinton}, L.~{Deng}, D.~{Yu}, G.~E. {Dahl}, A.~{Mohamed}, N.~{Jaitly},
  A.~{Senior}, V.~{Vanhoucke}, P.~{Nguyen}, T.~N. {Sainath}, and
  B.~{Kingsbury}, ``Deep neural networks for acoustic modeling in speech
  recognition: The shared views of four research groups,'' \emph{IEEE Signal
  Processing Magazine}, vol.~29, 2012.

\bibitem{Povey11-kaldi}
D.~Povey, A.~Ghoshal, G.~Boulianne, L.~Burget, O.~Glembek, N.~Goel,
  M.~Hannemann, P.~Motlicek, Y.~Qian, P.~Schwarz, J.~Silovsky, G.~Stemmer, and
  K.~Vesely, ``The kaldi speech recognition toolkit,'' in \emph{ASRU}, 2011.

\bibitem{Collobert16-wav2letter}
R.~{Collobert}, C.~{Puhrsch}, and G.~{Synnaeve}, ``Wav2letter: an end-to-end
  convnet-based speech recognition system,'' \emph{CoRR}, vol. abs/1609.03193,
  2016.

\bibitem{Liptchinsky17-letterbased}
V.~{Liptchinsky}, G.~{Synnaeve}, and R.~{Collobert}, ``Letter-based speech
  recognition with gated convnets,'' \emph{CoRR}, vol. abs/1712.09444, 2017.

\bibitem{Likhomanenko19-whoneedswords}
T.~{Likhomanenko}, G.~{Synnaeve}, and R.~{Collobert}, ``Who needs words?
  lexicon-free speech recognition,'' \emph{CoRR}, vol. abs/1904.04479, 2019.

\bibitem{Ashihara19-whspdetec}
T.~Ashihara, Y.~Shinohara, H.~Sato, T.~Moriya, K.~Matsui, T.~Fukutomi,
  Y.~Yamaguchi, and Y.~Aono, ``Neural whispered speech detection with
  imbalanced learning,'' in \emph{INTERSPEECH}, 2019.

\bibitem{Cotescu19-whspvc}
M.~Cotescu, T.~Drugman, G.~Huybrechts, J.~Lorenzo-Trueba, and A.~Moinet,
  ``Voice conversion for whispered speech synthesis,'' \emph{IEEE Signal
  Processing Letters}, vol.~27, 2019.

\bibitem{Salvador07-fastdtw}
S.~Salvador and P.~Chan, ``Toward accurate dynamic time warping in linear time
  and space,'' \emph{Intelligent Data Analysis}, vol.~11, 2007.

\bibitem{Sundermeyer15-RNNLM}
M.~Sundermeyer, H.~Ney, and R.~Schlüter, ``From feedforward to recurrent lstm
  neural networks for language modeling,'' \emph{IEEE/ACM Transactions on
  Audio, Speech, and Language Processing}, vol.~23, 2015.

\end{thebibliography}

\end{document}